\documentclass{article}

\usepackage[nonatbib,preprint]{neurips_2019}

\usepackage[utf8]{inputenc} 
\usepackage[T1]{fontenc}    
\usepackage{hyperref}       
\usepackage{url}            
\usepackage{booktabs}       
\usepackage{amsfonts}       
\usepackage{nicefrac}       
\usepackage{microtype}      
\usepackage{amsmath}
\usepackage[pdftex]{graphicx}
\usepackage[numbers]{natbib}

\title{Set Flow: A Permutation Invariant Normalizing Flow}

\author{Kashif Rasul \\ 
Zalando Research\\
Mühlenstraße 25, 10243 Berlin, Germany\\
\texttt{kashif.rasul@zalando.de}
\And 
Ingmar Schuster \\
Zalando Research\\
Mühlenstraße 25, 10243 Berlin, Germany\\
\texttt{ingmar.schuster@zalando.de}
\And
Roland Vollgraf\\
Zalando Research\\
Mühlenstraße 25, 10243 Berlin, Germany\\
\texttt{roland.vollgraf@zalando.de}
\And
Urs Bergmann \\
Zalando Research\\
Mühlenstraße 25, 10243 Berlin, Germany\\
\texttt{urs.bergmann@zalando.de}
}

\begin{document}

\maketitle

\begin{abstract}
    We present a generative model that is defined on \emph{finite sets} of exchangeable, potentially high dimensional, data. As the architecture is an extension of Real NVPs, it inherits all its favorable properties, such as being invertible and allowing for exact log-likelihood evaluation. We show that this architecture is able to learn finite non-i.i.d. set data distributions, learn statistical dependencies between entities of the set and is able to train and sample with variable set sizes in a computationally efficient manner. Experiments on 3D point clouds show state-of-the art likelihoods.
\end{abstract}



\section{Introduction}
Most of machine learning research concerns itself with either modeling independent and identically distributed (i.i.d.) data, or a full joint probability over a number of variables is modeled, i.e. $p(\mathbf{x}_1, ..., \mathbf{x}_s)$. 
However, some data is conceptually best represented as a finite unordered set: e.g. point clouds of objects, voice data from a given speaker, or documents as bag-of-words. This is why there has been growing interest in set modeling typically via composition of elementwise operations with permutation invariant reduction operations such as averaging as in~\cite{NIPS2017_6931} or taking the maximum as in~\cite{Qi2017PointNetDL} which introduces a bottleneck in what information about the set can be extracted.    


Formally, any finite joint probability distribution over $s$  exchangeable random variables $\mathbf{x}_i$ (called ``entities'' from now on) must fulfill the following requirement for all $s!$ permutations $\pi$:
\begin{equation}
    p(\mathbf{x}_1, \ldots, \mathbf{x}_s) = p(\mathbf{x}_{\pi(1)}, \ldots, \mathbf{x}_{\pi(s)}).
\end{equation}
It has been shown that finite exchangeable distributions can be written as a \textit{signed mixture} of i.i.d. distributions~\cite{Kerns2006}. Note that this differs from de Finetti's theorem, which is a similar statement for exchangeable processes, i.e. infinite sequences of random variables, and states that in this case the distribution is a mixture of i.i.d. processes: $p(\mathbf{x}_1, .., \mathbf{x}_s) = \int p(\theta) \prod_{i=1}^s p(\mathbf{x}_i|\theta)\, d\theta$, with $p(\theta)$ a probability distribution. For illustration of this difference, take a distribution of two exchangeable random variables that sum up to a fixed number $N$ (here commutativity ensures exchangeability)---in this case sampling of the two numbers cannot be written as conditionally independent given a $\theta$. Recent generative models build on top of de Finetti's results \cite{NIPS2018_7949, DBLP:journals/corr/abs-1902-01967}, and hence assume an underlying infinite sequence of exchangeable variables.

In this work we present an architecture to explicitly model finite exchangeable data. In other words, we are concerned with data comprised of sets $X$ which are i.i.d. sampled from our underlying data, but each set $X = \{\mathbf{x}_i\}_{i=1}^{s}$ is a finite exchangeable set, which can be non-i.i.d. data samples in some arbitrary order. We develop a density estimation model that is permutation invariant and is able to model dependencies between the entities in this setting. We call the resulting architecture \emph{Set Flow}, as it builds on ideas of normalizing flows, in particular compositions of bijections like Real NVP~\cite{45819}, and combines these ideas with set models~\cite{NIPS2017_6931}.


The paper is structured as follows. In Section 2 we review background concepts and Section 3 has related work. Section 4  describes our model and how it is trained. In Section 5 we present experiments on synthetic and real data, and finally conclude in Section 6 with a brief summary and discussion of future directions.

\section{Background}

\subsection{Sets}

One straight-forward approach to generate a set function is to treat the input as a sequence and train an RNN, but augmented with all possible input permutations, in the hopes that the RNN will become invariant to the input order. This approach might be robust to small sequences but for set sizes in the thousands it becomes hard to scale. Also, as described in~\cite{44871}, the order of the sequences does matter and cannot be discarded.

A recently proposed neural network method, which is invariant to the order if its inputs, is the Deep Set architecture~\cite{NIPS2017_6931}. The key idea of this approach is to map each input to a learned feature representation, on which a pooling operation is performed (e.g. a sum), which then is passed through another function. With $\mathcal{X}$ being the set of all sets, $X \in \mathcal{X}$ being a set, the deep set function $f: \mathcal{X} \mapsto \mathbb{R}^S$ can be written as $f(X) = \rho \left( \sum_{\mathbf{x} \in X} \phi(\mathbf{x}) \right)$, where $\phi \colon \mathbb{R}^D \mapsto \mathbb{R}^K$ and $\rho \colon \mathbb{R}^K \mapsto \mathbb{R}^S$ are chosen as a neural networks.


Recent methods like Janossy pooling~\cite{murphy2018janossy} expresses a permutation invariant function as the average of a permutation variant function applied to all reorderings of the input sequence which allows the layer to leverage complicated permutation variant functions to construct permutation invariant ones. This is computationally demanding, but can be done in a tractable fashion via approximation of the ordering or via random permutations. One can also train a permutation optimization module that learns a canonical ordering~\cite{zhang2019permoptim} to permute a set and then use it in a permutation invariant fashion, typically by processing it via an RNN.

\subsection{Density Estimation via Normalizing Flows}

Real NVP~\cite{45819} is a type of normalizing flow~\cite{tabak} where densities in the input space $\mathcal{X}=\mathbb{R}^{D}$ are transformed into some simple distribution space $\mathcal{Z}=\mathbb{R}^{D}$, like an isotropic Gaussian, via $f \colon \mathcal{X} \mapsto \mathcal{Z}$, which is composed of stacks of bijections or invertible mappings, with the property that the inverse $\mathbf{x} = f^{-1}(\mathbf{z})$ is easy to evaluate and computing the Jacobian determinant takes $O(D)$ time. Due to the change of variables formula we can evaluate $p_{\mathcal{X}}(\mathbf{x})$ via the  Gaussian by
\begin{equation}
p_{\mathcal{X}}(\mathbf{x}) = p_{\mathcal{Z}}(\mathbf{z}) \left| \det \left( \frac{\partial f(\mathbf{x})}{\partial \mathbf{x}} \right)\right|.
\label{eq:variable_change}
\end{equation}

The bijection introduced by Real NVP called the \emph{Coupling Layer} satisfies the above two properties. It leaves part of its inputs unchanged and  transforms the other part via functions of the un-transformed variables
\[
\begin{cases}
    \mathbf{y}^{1:d} = \mathbf{x}^{1:d} \\
    \mathbf{y}^{d+1:D} =  \mathbf{x}^{d+1:D} \odot \exp(s( \mathbf{x}^{1:d})) + t( \mathbf{x}^{1:d}),
\end{cases}
\]
where $\odot$ is an element wise product, $s$ is a scaling and $t$ a translation function from $\mathbb{R}^{d} \mapsto \mathbb{R}^{D-d}$, given by  neural networks. To model a complex nonlinear density map $f(\mathbf{x})$, a number of coupling layers $\mathcal{X} \mapsto \mathcal{Y}_1 \mapsto \cdots \mapsto \mathcal{Y}_{K-1} \mapsto \mathcal{Z}$ are composed together, while alternating the dimensions which are unchanged and transformed. Via the change of variables formula the probability density function (PDF) of the flow given a data point can be written as
\begin{equation}\label{Real NVP-logp}
\log p_{\mathcal{X}}(\mathbf{x}) = \log p_{\mathcal{Z}}(\mathbf{z}) + \log | \det(\partial \mathbf{z}/ \partial\mathbf{x})| = \log p_{\mathcal{Z}}(\mathbf{z}) +  \sum_{i=1}^{K} \log | \det(\partial \mathbf{y}_{i}/ \partial\mathbf{y}_{i-1})|.    
\end{equation}

Note that the Jacobian for the Real NVP is a block-triangular matrix and thus the log-determinant simply becomes
\begin{equation}\label{Real NVP-logdet}
    \log | \det(\partial \mathbf{y}_{i}/ \partial\mathbf{y}_{i-1})| = \mathtt{sum}(\log|\mathtt{diag}( \exp (s(\mathbf{y}_{i-1}))|),
\end{equation}
where $\mathtt{sum}()$ is the sum over all the vector elements, $\log()$ is the element-wise logarithm and $\mathtt{diag}()$ is the diagonal of the Jacobian. This model, parameterized by the weights of the scaling and translation neural networks $\theta$, is then trained via  stochastic gradient descent (SGD) on training data points where for each batch $\mathcal{D}$ the log likelihood (\ref{Real NVP-logp}) as given by
\[
\mathcal{L} = \frac{1}{|\mathcal{D}|} \sum_{\mathbf{x}\in\mathcal{D}}  \log p_{\mathcal{X}}(\mathbf{x}; \theta),
\]
is maximized. One can trivially condition the PDF on some additional information  $\mathbf{h} \in \mathbb{R}^H$ to model  $p_{\mathcal{X}}(\mathbf{x} | \mathbf{h})$ by concatenating $\mathbf{h}$ to the inputs of the scaling and translation function approximators, i.e. $s(\mathtt{concat}(\mathbf{x}^{1:d}, \mathbf{h}))$ and $t(\mathtt{concat}(\mathbf{x}^{1:d}, \mathbf{h}))$ which are modified to map $\mathbb{R}^{d+H} \mapsto \mathbb{R}^{D-d}$. This does not change the log-determinant of the coupling layers given by (\ref{Real NVP-logdet}).

In practice  Batch Normalization~\cite{Ioffe:2015:BNA:3045118.3045167} is applied, as a bijection, to outputs of coupling layers to stabilize training of normalizing flow. This bijection implements the normalization procedure using  a weighted average of a moving average of the layer's mean and standard deviation values, which are different depending if we are training or doing inference.

\section{Related Work}

The Real NVP approach can be generalized as in the Masked Autoregressive Flow (MAF)~\cite{Papamakarios:2017:maf} which models the random numbers used in each stack to generate data. Glow~\cite{NIPS2018_8224} augments Real NVP by the addition of a reversible $1 \times 1$ convolution, as well as removing other components and thus simplifying the overall architecture to obtain qualitatively better samples for high dimensional data like images.

The BRUNO model~\cite{NIPS2018_7949} performs exact Bayesian inference on sets of data such that the joint distribution over observations is permutation invariant in an autoregressive fashion, in that new samples can be generated conditional on previous ones and a stream of new data points can be easily incorporated at test time.
This is easily possible for our method as well, where the network architecture is considerably simple as it only draws upon ideas from normalizing flows.
BRUNO, on the other hand, makes use of Student-$t$ processes, i.e. Bayesian models of real-valued functions admitting closed form marginal likelihood and posterior predictive expressions \cite{shah2014student}.
The main issue with this building block is that inference typically scales cubically in the number of data points, although the Woodbury matrix inversion lemma can be used to alleviate this issue for the streaming data setting.

Similar to BRUNO, the PILET model~\cite{DBLP:journals/corr/abs-1902-01967} utilizes an autoregressive model, build upon normalizing flow ideas instead of Student-$t$-processes~\cite{pmlr-v80-oliva18a}. This is combined with a permutation equivariant function to capture interdependence of entities in a set while maintaining exchangeability. They extend their method to make use of a latent code in an exchangeable variational autoencoder framework called PILET-VAE. Note both BRUNO and PILET transform base distributions by applying bijections to entity dimension. 

Bayesian Sets~\cite{NIPS2005_2817} also models exchangeable sets of binary features but it is not reversible so does not allow sampling from it.

\section{Set Flow}
In order to make a model invariant to input permutations, one can try to sort the input into some canonical order. While sorting is a very simple solution,  for high dimensional points the ordering is in general not stable with respect to the point perturbations and thus does not fully resolve the  issue. This makes it hard for a model to learn a consistent mapping even if we constrain the model to have the same set size.

We propose a normalizing flow architecture called Set Flow that in each stack transforms each entity of the set via a shared global Gaussian noise vector, and then this  noise vector gets transformed via a symmetric function of all the transformed elements of the set, for example via a Deep Set~\cite{NIPS2017_6931} layer. 

\begin{figure}
  \centering
  \includegraphics[width=0.7\linewidth]{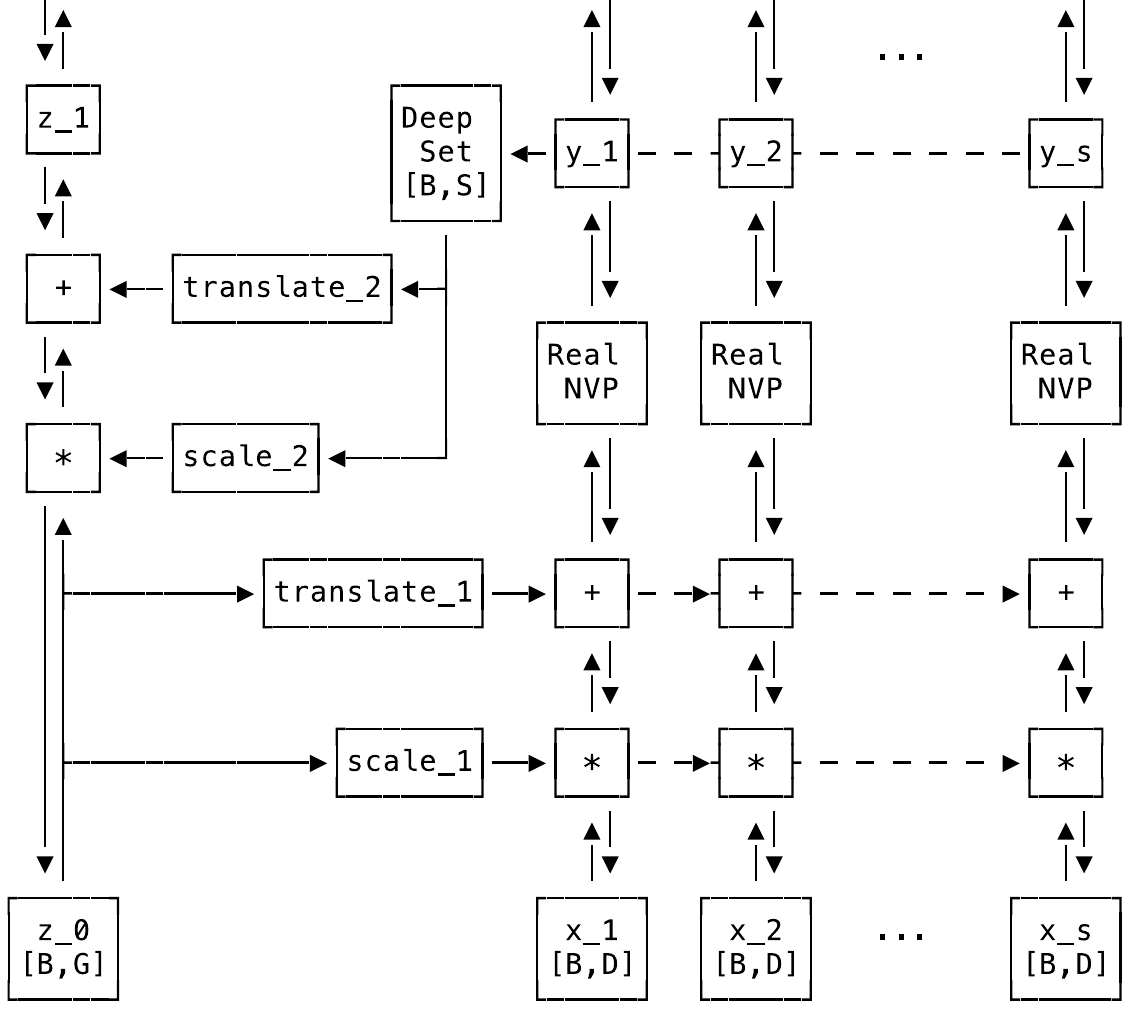}
  \caption{Schematic of a single Set Flow stack where a set of entities $\{ \mathbf{x}_1, \ldots \mathbf{x}_s\}$, where $\mathbf{x}_i \in \mathbb{R}^D$ and a global Gaussian noise vector $\mathbf{z}_0 \sim \mathcal{N}(0, I) \in \mathbb{R}^G$, are transformed $ (\mathbf{z}_0, \{\mathbf{x}_1, \ldots, \mathbf{x}_s\}) \mapsto (\mathbf{z}_1, \{\mathbf{y}_1, \ldots, \mathbf{y}_s\})$ via (\ref{eq:set-flow}). See text for detailed description.}
  \label{fig:set-flow}
\end{figure}

Figure~\ref{fig:set-flow} shows a single Set Flow stack, which takes its input from layer $k=0$ to the next stack $k=1$. The block takes a set of entities $\{ \mathbf{x}_1, \ldots \mathbf{x}_s\}$ where $\mathbf{x}_i \in \mathbb{R}^D$, and a global Gaussian noise vector $\mathbf{z}_0 \sim \mathcal{N}(0, I) \in \mathbb{R}^G$ and transforms it to $ (\mathbf{z}_0, \{\mathbf{x}_1, \ldots, \mathbf{x}_s\}) \mapsto (\mathbf{z}_1, \{\mathbf{y}_1, \ldots, \mathbf{y}_s\})$  given by:
\begin{equation}
\begin{cases}
    \hat{\mathbf{y}}_i =  \mathbf{x}_i \odot \exp(s_1^{(0)}( \mathbf{z}_0)) + t_1^{(0)}( \mathbf{z}_0) \quad \mathrm{for} \quad i=1,\ldots,s\\
    \hat{\mathbf{y}}_i \mapsto \mathbf{y}_i  \quad  \mathrm{via\ RealNVP}^{(0)}  \mathrm{\ for}  \quad  i=1,\ldots,s \quad \mathrm{if} \quad D > 1 \\  
    \mathbf{z}_1 =  \mathbf{z}_0 \odot \exp(s_2^{(0)}( f^{(0)}(\mathbf{y}_1, \ldots,\mathbf{y}_s ))) + t_2^{(0)}(f^{(0)}(\mathbf{y}_1, \ldots, \mathbf{y}_s))
\end{cases}
\label{eq:set-flow}
\end{equation}
where $f^{(k)}$ is a permutation invariant function given via a Deep Set, $t_i^{(k)}$ and $s_i^{(k)}$ are deep neural networks function approximators and $\mathrm{RealNVP}^{(k)}$ is a standard Real NVP---all of these functions are layer $k$ specific and do not share weights across layers. By stacking $K$  such Set-Coupling layers we arrive at our \emph{Set Flow} model.  As one can see from the construction this mapping is permutation equivariant due to the Deep Set layer and invertable via the bijections.

The inverse transformation starts by sampling a global noise vector $\mathbf{z}_{K-1} \sim \mathcal{N}(0, I) \in \mathbb{R}^G$ as well as a set of the desired number of Gaussian  sample entities and going through the flow model in reverse (or from the top to bottom in Figure~\ref{fig:set-flow}).

As in the Real NVP we can also condition this model via some $\mathbf{h} \in \mathbb{R}^H$ for each set of entities $\{\mathbf{x}_i\}_{i=1}^{s}$ by the following modification in (\ref{eq:set-flow}):
\[
\hat{\mathbf{y}}_i =  \mathbf{x}_i \odot \exp(s_1^{(0)}( \mathtt{concat}(\mathbf{z}_0, \mathbf{h}))) + t_1^{(0)}( \mathtt{concat}(\mathbf{z}_0, \mathbf{h}))
\]
to obtain a set-conditioned model, for example when the entities of a set come from a particular category.

\subsection{Training}

We train the model by sampling batches where for each batch  $\mathcal{D}$ the size of the set $s$ is fixed, and construct $|\mathcal{D}|$ sets where each set has $s$ entities as well as a global noise vector. 
We use Adam~\cite{kingma:adam} with standard parameters, to maximize the log likelihood:
\begin{align}
    \mathcal{L} &= \mathcal{L}_{\mathcal{X}} + \mathcal{L}_{\mathcal{N}} \nonumber \\
     &= \sum_{i=1}^{s} \log p_{\mathcal{X}} (\mathbf{x}_i; \theta) + \log p_{\mathcal{N}}(\mathbf{z}_0; \theta),
     \label{eq:SetFlowLikelihood}
\end{align}
where for each term above (\ref{eq:variable_change}) is employed to explicitly evaluate the likelihoods and  calculate their derivatives, with respect to $\theta$ which denotes all parameters of the Set Flow model.

Note that we choose $\mathbf{z}_0 \sim \mathcal{N}(0, I) \in \mathbb{R}^G$ in all our experiments. As we're interested in the likelihoods of the sets we hence subtract $G$ times the entropy of a Gaussian (the maximum likelihood solution of the global variables) with variance $1$ from the calculated likelihoods of (\ref{eq:SetFlowLikelihood}) when reporting the test set likelihoods.



\section{Experiments}
Our first goal in the experiments is to demonstrate and analyze the ability of the proposed model to capture non-i.i.d. dependencies within finite sets. In a second set of experiments, we show that the model scales to much larger and complex datasets by learning 3D point clouds.

\subsection{Generation of Non-i.i.d. Exchangeable Data Sets}
In order to best understand the ability of the model to capture dependencies of entities, we generate a toy dataset of finite sets with a non-i.i.d. structure: equidistant 2D points on circles with varying radius and position. The generative process of each set is given as follows: first, the center position $x,y$, radius $r$ and a rotation offset $\phi$ is sampled uniformly as $x,y \sim \mathcal{U}(-10,10)$, $r \sim \mathcal{U}(0.5,3)$ and $\phi \sim \mathcal{U}(0,2\pi)$. Then $N$ points are generated with coordinates $x_i = x + (r + \Delta r_i) \; \mathrm{cos}(\psi_i)$ and $y_i = y + (r + \Delta r_i) \; \mathrm{sin}(\psi_i)$, where $\psi_i = \phi + 2\pi i/N + \Delta \psi_i$, with independent radial noise $\Delta r_i \sim \mathcal{N}(0,0.1)$ and angular noise $\Delta \psi_i \sim \mathcal{N}(0,0.3)$.

\begin{figure}
  \centering
  \includegraphics[width=0.261\linewidth]{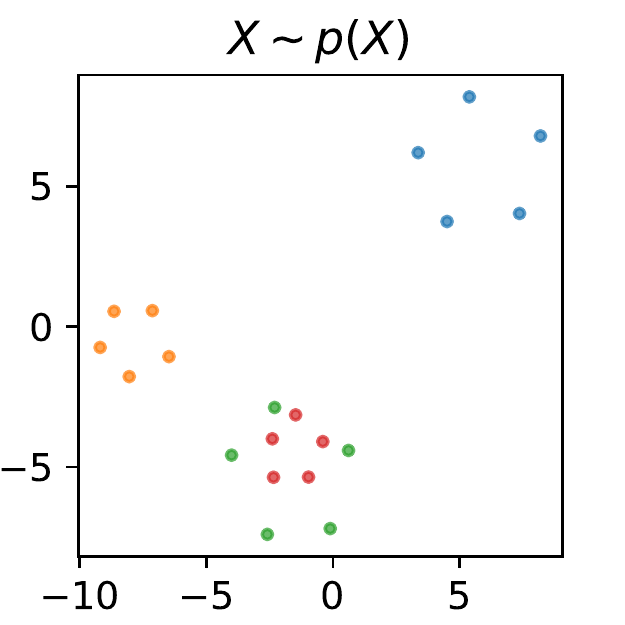}
  \hspace{-0.3cm}
  \includegraphics[width=0.261\linewidth]{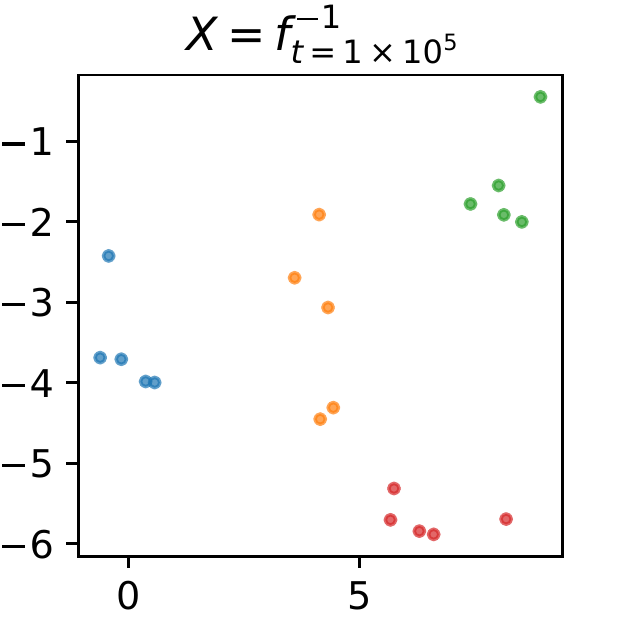}
  \hspace{-0.45cm}
  \includegraphics[width=0.261\linewidth]{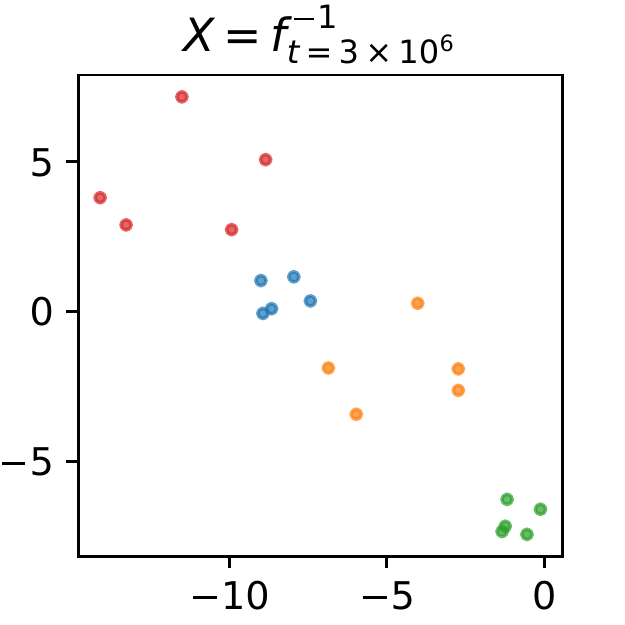}
  \hspace{-0.45cm}
  \includegraphics[width=0.261\linewidth]{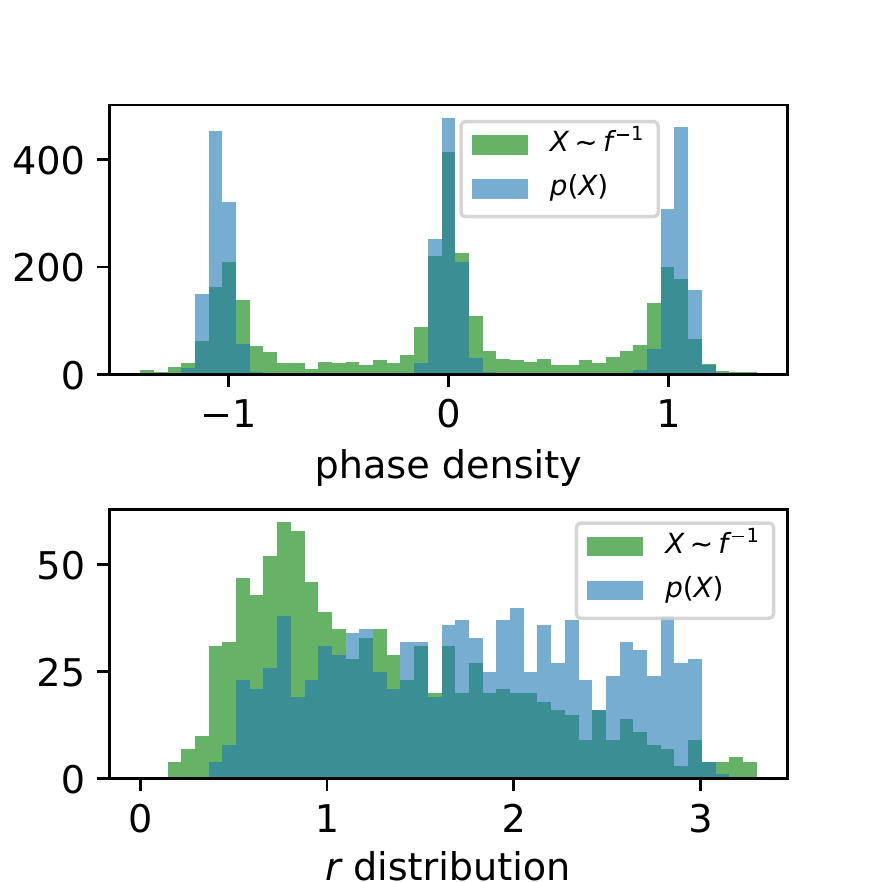}
  \caption{Non-i.i.d. analysis. The leftmost subfigure shows samples of sets with $N=5$ 2D equidistant entities drawn on circles with random positions and radii. The model initially captures the global position variance in the data (second from left at $t=10^5$ samples) and later in learning captures the non-i.i.d. equidistance property on the circles (at $t=3\times10^6$, second from right). The rightmost subfigure shows that the model nicely captures equidistant phases on the sampled circles, but has a bias towards smaller circles than in the original dataset.}  \label{fig:noniid}
\end{figure}

Figure~\ref{fig:noniid}~(left) shows sample sets with a size $N=5$ drawn from this generative model---colors indicate set membership. For the experiment, we trained a model on uniformly random sampled set sizes in $\{3,4,5,6\}$, where each minibatch of $16$ sets contained the same set sizes. The second subfigure from left shows that after $10^5$ set samples, the model groups elements of sets together in clusters, but fails to produce discernible circles with equidistant points on them. After $3\times10^6$ set samples, the model reproduces the dataset more faithfully, as can be seen in the second to right Figure. The rightmost subfigure in Figure~\ref{fig:noniid} (top) shows the distribution of inferred phases from fitted circles to sets of size $N=3$ (the mean phase across the set is subtracted for alignment). It can be seen that the model (green) nicely captures the equidistant peaks, similar to the original data (blue). Note that this implies that the model captured the generative process of the finite set---otherwise there would be more mass in between the peaks. The variance, however, is larger than in the ground truth phases. Similarly, the model has a bias towards smaller circles---as can be seen in the distribution of inferred radii.

\subsection{3D Point Cloud Experiments}

\begin{figure}
  \centering
 \fbox{\includegraphics[trim={8mm 8mm 8mm 8mm}, clip, width=0.47\linewidth]{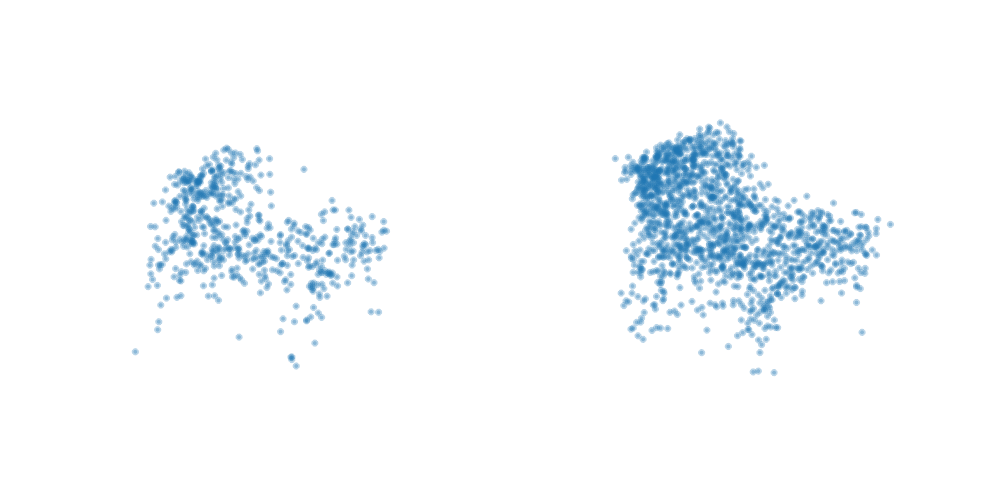}} \fbox{\includegraphics[trim={8mm 8mm 8mm 8mm}, clip, width=0.47\linewidth]{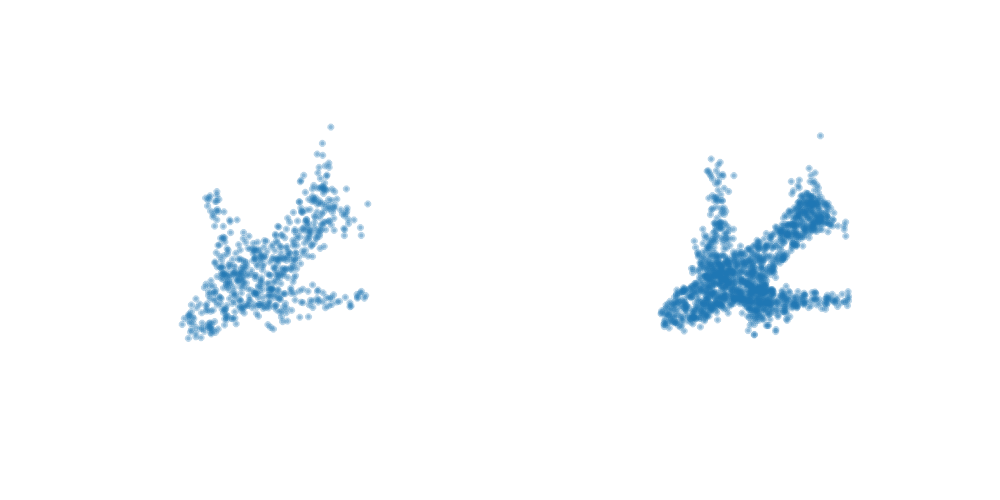}}
 \caption{Samples with different number of entities from Set Flow trained on chairs and airplane models separately.}
  \label{fig:chair-plane-samples}
\end{figure}

We train Set Flow from point clouds of Airplane and Chair classes of the ModelNet40~\cite{CVPR15_Wu} dataset, where we sample $s=1,000$ random points from a point cloud of 10,000 for each model to construct a set for the chosen class. We split the model files into a training and test set via an 80\% split. We train the model on two individual classes: airplane and chair separately and report the mean test likelihoods in Table~\ref{tb:modelnet40}. We also show some sample generated point clouds in Figure~\ref{fig:chair-plane-samples} for different set sizes. 

\begin{figure}
  \centering
  \fbox{\includegraphics[trim={8mm 8mm 8mm 8mm}, clip, width=\linewidth]{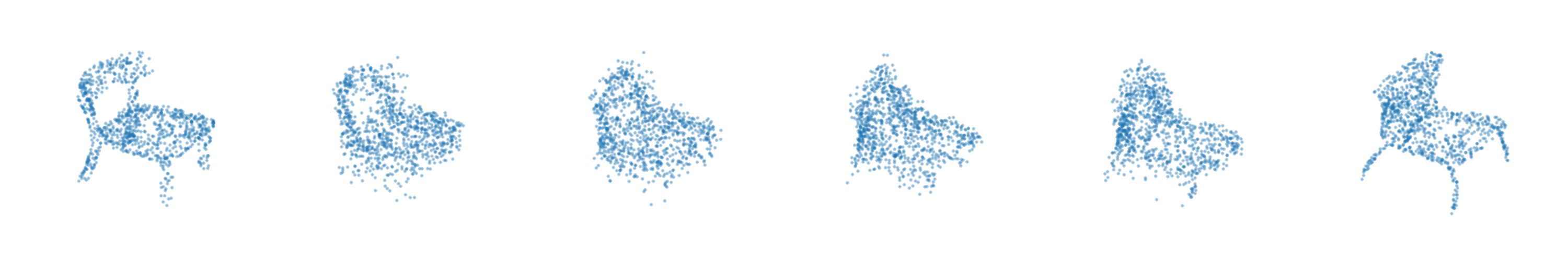}}\\
  \fbox{\includegraphics[trim={8mm 8mm 8mm 8mm}, clip, width=\linewidth]{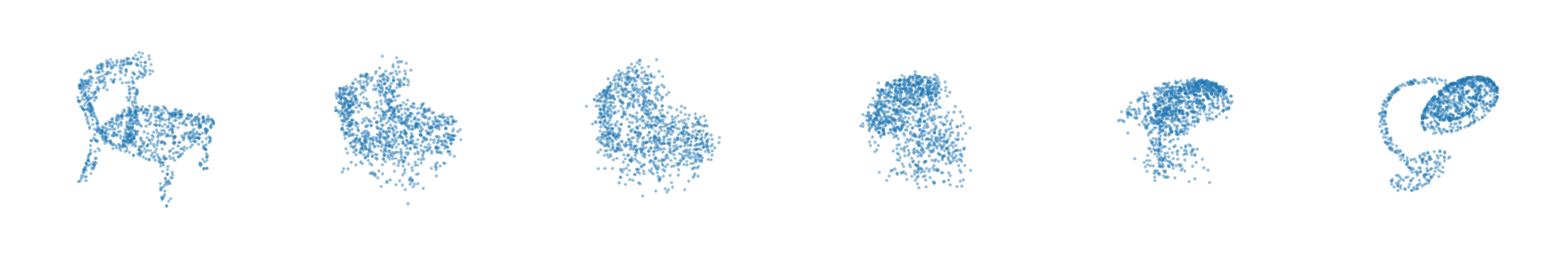}}\\
  \fbox{\includegraphics[trim={8mm 8mm 8mm 8mm}, clip, width=\linewidth]{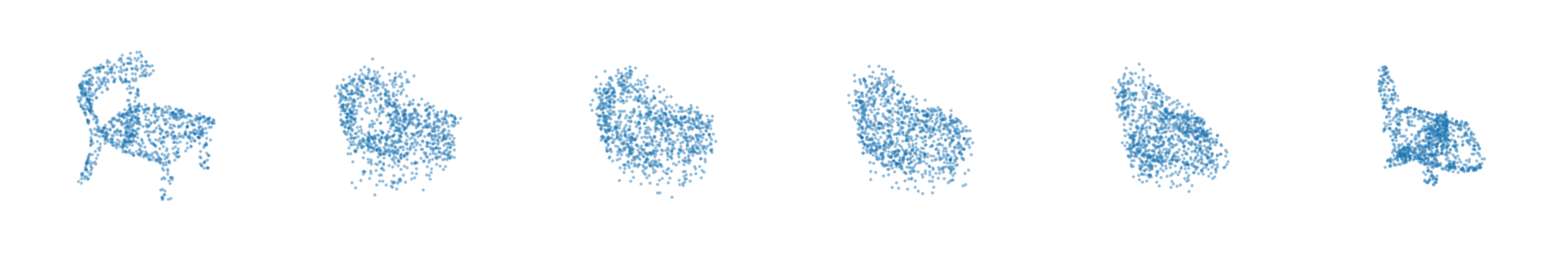}}
  \caption{Generated samples from interpolating the noise vectors obtained from two models (left-most and right-most) from Set Flow trained on chairs, lamp and airplane models together.}
  \label{fig:chairs-inter}
\end{figure}

We also train the model on three classes (airplane, chair and lamp) together and then given two sets we obtain the noise vectors by passing the sets through our model. We can then linearly interpolate between these two sets and generate samples by passing the interpolated noise, both global and for the entities of the set, backwards. Figure~\ref{fig:chairs-inter} shows the results of this experiment for a chair to another chair, chair to a lamp and chair to an airplane.

Finally, we train the model on all 40 classes both without supplying the class labels and with class labels via a set class embedding vector $\mathbf{h}\in \mathbb{R}^H$. We report the mean test log-likelihoods over each entity in the set in Table~\ref{tb:modelnet40} together with results from other methods. 

We have implemented all the experiments in PyTorch~\cite{paszke2017automatic} and will make the code available after the review process here\footnote{\url{https://www.github.com/xxx/xxx}}. We used the following hyperparameters: batch size $|\mathcal{D}|=16$, global noise vector dimension $G=90$, size of Deep Set pooling output $S =100$, size of conditioning embedding vector $H=10$, number of Set Flow stacks $K=6$, number of random entities in a set $s=1,000$ and a learning rate of $5 \times 10^{-4}$ for all our experiments.

\begin{table}
\caption{Mean  test log-likelihoods for ModelNet40~\cite{CVPR15_Wu} dataset models with two times standard error for our method.}
 \label{tb:modelnet40}
  \centering
 \begin{tabular}{lcccc}
  \toprule

  Model     & Airplane     & Chair & All & All with labels\\
  \midrule
  PILET-VAE & $\textbf{4.08}$ & $\textbf{2.03}$ & \textbf{2.13} & \textbf{2.29} \\
  PILET & $3.75$ & $1.47$  & $1.58$ & $1.94$ \\
  BRUNO     & $2.6$ & $0.79$ & $0.75$ & $0.64$     \\
  Set Flow     & \textbf{4.1} $\pm 0.35$       &  \textbf{2.045} $\pm 0.25$   &  \textbf{2.143} $\pm 0.318$ &  \textbf{2.311} $\pm 0.298$\\
   \bottomrule
 \end{tabular}
\end{table}

\section{Discussion and Conclusions}

We have introduced a simple generative architecture for learning and sampling from exchangeable data of finite sets via a normalizing flow architecture using permutation invariant functions like, for example, Deep Sets. As shown in the experiments our model captures dependencies between entities of a set in a computationally feasible manner. We demonstrated the capability of the model to capture finite exchange invariant generative processes on toy data. We also demonstrated state-of-the art performance for generative modeling of 3D point clouds.
In principle, the propose model can be applied to higher dimensional data points, like for example sets of images e.g. in an outfit.

In future work we will further explore alternative architectures of these models, utilize them to learn on sets of images and experiment to see if these methods can be used to learn correlations in time series data across a large number of entities.

\bibliographystyle{abbrvnat}
\bibliography{references}


\end{document}